\documentclass[a4paper, 10pt, conference]{ieeeconf}      

\IEEEoverridecommandlockouts                              

\usepackage{graphics} 
\usepackage{epsfig} 
\usepackage{mathptmx} 
\usepackage{times} 
\usepackage{amsmath} 
\usepackage{amssymb}  
\usepackage{cite}
\usepackage[T1]{fontenc}
\usepackage{xcolor}
\usepackage{graphicx}
\usepackage[all]{nowidow}
\usepackage{tikz,hyperref}
\usepackage{microtype}
\usepackage{makecell}
\usepackage{comment}
\usepackage{multirow}
\usepackage[caption=false,font=normalsize,labelfont=normalfont,textfont=normalfont]{subfig}
\usepackage{arydshln}
\title{\LARGE \bf Predicting and Analyzing Pedestrian Crossing Behavior at \\ Unsignalized Crossings
}

\author{Chi Zhang$^{1}$,
Janis Sprenger$^{2}$,
Zhongjun Ni$^{3}$,
and Christian Berger$^{1}$
\thanks{*This research is partially funded by the research project ``SHAPE-IT – Supporting the Interaction of Humans and Automated Vehicles: Preparing for the Environment of Tomorrow’’ which has received funding from the European Union’s Horizon 2020 research and innovation programme under the Marie Skłodowska-Curie grant agreement 860410, and partially by the German Ministry for Research and Education (BMBF) in the project project MOMENTUM (01IW22001).}
\thanks{$^{1}$Department of Computer Science and Engineering, University of Gothenburg, Sweden.
        {\tt\small \{chi.zhang, christian.berger\}@gu.se}}%
\thanks{$^{2}$German Research Center for Artificial Intelligence (DFKI), Saarland Informatics Campus, Germany.}%
\thanks{$^{3}$Department of Science and Technology, Link{\"o}ping University, Campus Norrk{\"o}ping, Sweden.}%
}

\begin{document}

\maketitle
\thispagestyle{empty}
\pagestyle{empty}

\begin{abstract}
Understanding and predicting pedestrian crossing behavior is essential for enhancing automated driving and improving driving safety. Predicting gap selection behavior and the use of zebra crossing enables driving systems to proactively respond and prevent potential conflicts. This task is particularly challenging at unsignalized crossings due to the ambiguous right of way, requiring pedestrians to constantly interact with vehicles and other pedestrians.
This study addresses these challenges by utilizing simulator data to investigate scenarios involving multiple vehicles and pedestrians. We propose and evaluate machine learning models to predict gap selection in non-zebra scenarios and zebra crossing usage in zebra scenarios. We investigate and discuss how pedestrians' behaviors are influenced by various factors, including pedestrian waiting time, walking speed, the number of unused gaps, the largest missed gap, and the influence of other pedestrians.
This research contributes to the evolution of intelligent vehicles by providing predictive models and valuable insights into pedestrian crossing behavior.

\end{abstract}

\section{Introduction}

The rapid evolution of automated driving and intelligent vehicles demands safe and intelligent pedestrian-vehicle interactions. Understanding and predicting pedestrian behavior is essential not only for enhancing the performance of automated driving systems but also for enabling human drivers to make safer decisions. A particularly challenging task in this domain is to predict crossing behavior at unsignalized crossings, due to the absence of signal displays and traffic lights introducing ambiguity of the right of way.

In the contemporary and foreseeable future, it is common for pedestrians to interact with a traffic flow of vehicles, either autonomous or human-driven, or a hybrid combination of both types, at unsignalized crossings. In such situations, the ability to understand and predict the time gap selected by a pedestrian for crossing the street is crucial. This understanding enables vehicles within the traffic flow to respond proactively.
Furthermore, contrary to the ideal situation where pedestrians consistently use the zebra crossings for safety, a study by D{\k{a}}browska-Loranc et al.~\cite{dkabrowska2021road} observed cases where pedestrians did not use zebra crossings. Hence, predicting pedestrians' intention to use zebra crossings in advance helps vehicles to avoid potential conflicts and collisions.

Predicting pedestrian behavior is challenging due to the simultaneous influence of multiple factors ~\cite{zhang2023pedestrian,Rasouli2019Autonomous}. Although previous studies have made significant strides on trajectory prediction~\cite{alahi2016social,gupta2018social,mohamed2020social,zhang2021social,zhang2022learning,zhang2023spatial} and crossing intention prediction~\cite{Rasouli2017they,Rasouli2019PIE,yang2021crossing}, these studies did not focus on unsignalized crossings. Additionally, they did not address pedestrians' gap selection behavior and their usage of zebra crossings. Therefore, our study focuses specifically on these two aspects.

Pedestrians constantly communicate and react to vehicles and other pedestrians while making crossing decisions. When investigating pedestrian-vehicle interactions, the study should be conducted safely without bringing pedestrians into hazardous situations and should avoid real pedestrian-vehicle conflicts. Hence, many researchers introduced simulator studies for investigating pedestrian interactions.
Kalantari et al.~\cite{Kalantari2022Who} and Zhang et al.~\cite{zhang2023cross} studied pedestrian-vehicle interactions at unsignalized crossings, but their studies were under a single-pedestrian encountering a single-vehicle scenario. However, real-world scenarios rarely involve a single pedestrian and vehicle only. Typically, a pedestrian interacts with multiple vehicles and other pedestrians.
Moreover, when predicting crossing behavior at zebra crossings, many studies (e.g.,~\cite{habibovic2018communicating,jayaraman2020analysis,Kalantari2022Who, zhang2023cross}) commonly assumed that pedestrians will consistently use the zebra crossings. This assumption overlooks the practical reality that pedestrians may initiate crossing from a distance to the zebra, allowing for the possibility that they may choose not to use it at all. Few studies explored whether pedestrians use the zebra crossing in such situations. 

In addressing these research gaps, we utilize data captured in a virtual reality simulator experiment~\cite{sprenger2023cross} to analyze and predict pedestrian crossing behavior. 
We investigate complex scenarios where pedestrians interact with a traffic flow of multiple vehicles. For non-zebra crossing scenarios, we predict what gaps pedestrians will select for crossing and explore the impact of group behavior. For zebra crossing scenarios, instead of assuming pedestrians start directly at the zebra when crossing, we consider that the pedestrians start from a distance to the zebra, reflecting real-world conditions.
We utilize machine learning models to predict gap selection and zebra crossing usage, and analyze the influence of interacting factors contributing to pedestrian crossing behavior.
We particularly focus on the following research questions:

\begin{enumerate}
    \item [RQ1] How predictable is the pedestrian crossing behavior including the gap selection and the use of zebra crossings, using machine learning models with the observed information before the event?
    \item [RQ2] Which factors are important in determining the crossing decision, and how do they influence the pedestrian crossing behavior?
\end{enumerate}

The main contributions of this study include:
\begin{enumerate}
    \item We propose and evaluate machine learning models that predict pedestrian gap selection behavior. The neural network model achieves the best mean absolute error at 1.07 seconds. We identify the most important features of each model and investigate the impact of various factors, including the number of unused car gaps, the largest missed car gap, pedestrian waiting time, pedestrian average walking speed, and group behavior.
    \item We propose and evaluate machine learning models that predict the use of zebra crossings. The neural networks capture the non-linearity of pedestrian behavior and get the best predicting accuracy at 94.27\% compared with other models. We identify the most important features of each model and investigate the impact of the number of unused car gaps and waiting time. We also look into the relationship between accepted gaps and the use of zebra crossings.
\end{enumerate}
\section{Related Work}

\subsection{Pedestrian Crossing Behavior Prediction}

Recent studies have made significant strides in predicting pedestrian crossing behavior using naturalistic data~\cite{Fang2018, Rasouli2017they, Chaabane2020, yang2021crossing, zhang2021pedestrian}.
Most of these studies employed deep learning networks, mainly focusing on the appearance or postures of pedestrians and their nearby environment. However, these models provided limited explicit consideration of the pedestrian-vehicle interaction.

Studies~\cite{Volz2015, Volz2016, Zhang2020Research} addressed pedestrian-vehicle interactions when predicting pedestrian behavior at zebra crossings.
V{\"o}lz et al.~\cite{Volz2015, Volz2016} integrated relative position and velocity between pedestrians and vehicles to predict pedestrian crossing behavior. Zhang et al.~\cite{Zhang2020Research} predicted pedestrian crossing intention considering interaction factors such as time to arrival and the speed and position of both the vehicle and the pedestrian.
However, these studies mainly focused on single-vehicle and single-pedestrian interactions, neglecting the complex dynamics introduced by multi-pedestrian and multi-vehicle scenarios that are prevalent in real-world situations. Moreover, the gap selection behavior of pedestrians has not been addressed in these predictions.

Simulator data is becoming popular for investigating pedestrian crossing behavior, especially when considering pedestrian-vehicle interaction. Ensuring the safety of pedestrians during such studies is important, and simulators can provide an effective way of collecting interaction data in a controlled and safe environment. Jayaraman et al.~\cite{jayaraman2020analysis} leveraged simulator data to analyze and predict pedestrians' gap acceptance behavior at zebra crossings, considering waiting time and vehicle speed. Zhang et al.~\cite{zhang2023cross} utilized distributed simulator data to study pedestrian crossing behavior when interacting with a single vehicle, and used time gap, pedestrian waiting time, age, and gender, as well as personal traits for prediction. However, these studies assumed that pedestrians would consistently use zebra crossings, neglecting instances where pedestrians may choose not to use them.

\subsection{Analysis on Pedestrian Crossing Behavior}

Previous research has studied pedestrian crossing behavior when interacting with vehicles exploring various factors.
Studies investigated the impact of pedestrian demographics such as age and gender~\cite{gorrini2018observation, Rasouli2019Autonomous, velasco2021will, cloutier2017outta, Kalantari2022Who} on crossing behavior. Studies on personality traits, as explored in studies~\cite{rosenbloom2006sensation, wang2022effect, Kalantari2022Who}, contributed to understanding the psychological aspects influencing pedestrian decisions during crossings. Vehicle-related factors such as vehicle speed were considered in studies~\cite{Volz2016, theofilatos2021cross, yannis2013pedestrian}.

Interaction factors are also considered important for crossing decisions. 
The time gap, a crucial feature in crossing decision modeling, was explored in studies highlighting its significance in pedestrian crossing behavior~\cite{gorrini2018observation, theofilatos2021cross, velasco2021will, Kalantari2022Who, zhang2023cross}. Pedestrian waiting time is another noteworthy dimension and was studied in~\cite{Kalantari2022Who, theofilatos2021cross, zhang2023cross, yannis2013pedestrian}. Furthermore, the influence of the presence of a zebra crossing on pedestrian choices was investigated and analyzed in studies~\cite{habibovic2018communicating, Kalantari2022Who, zhang2023cross}.

However, existing studies primarily employ these factors to model the relationship with crossing decisions without predicting specific outcomes. For instance, Velasco et al.~\cite{velasco2021will} found that pedestrians are more likely to cross at larger gaps, while Gorrini et al.~\cite{gorrini2018observation} and Sprenger et al.~\cite{sprenger2023cross} calculated accepted gaps from a statistical perspective. Although Yannis et al.~\cite{yannis2013pedestrian} modeled traffic gaps and crossing decisions at mid-block crossings, considering factors such as waiting time, vehicle speed, and group behavior, their approach was limited to linear relationships with these factors, and they did not propose predictive models.
Few studies delve into predicting the specific gaps pedestrians would select and accept for crossing.
Moreover, when considering the impact of zebra crossings, existing studies often assume pedestrians start directly from the zebra and will invariably use it, overlooking the possibility that pedestrians may initiate their crossing from a distance to the zebra and may not use it.

To address the existing research gaps, we consider pedestrians' interactions with multi-vehicles. For non-zebra crossing scenarios, we predict the time gap pedestrians will select and accept when interacting with a flow of vehicle traffic. Factors such as waiting time, pedestrian walking speed, and the car gaps that were missed for crossing are analyzed. We also explore and analyze the influence of group behavior. For zebra crossings, we predict if pedestrians will use the zebra crossing, and analyze the factors that influence their choices.

\section{Methodology}

\begin{figure*}
    \centering
    \subfloat[Baseline environment and virtual pedestrians]{\includegraphics[page=6,width=0.47\textwidth,trim=20 0 20 0,clip]{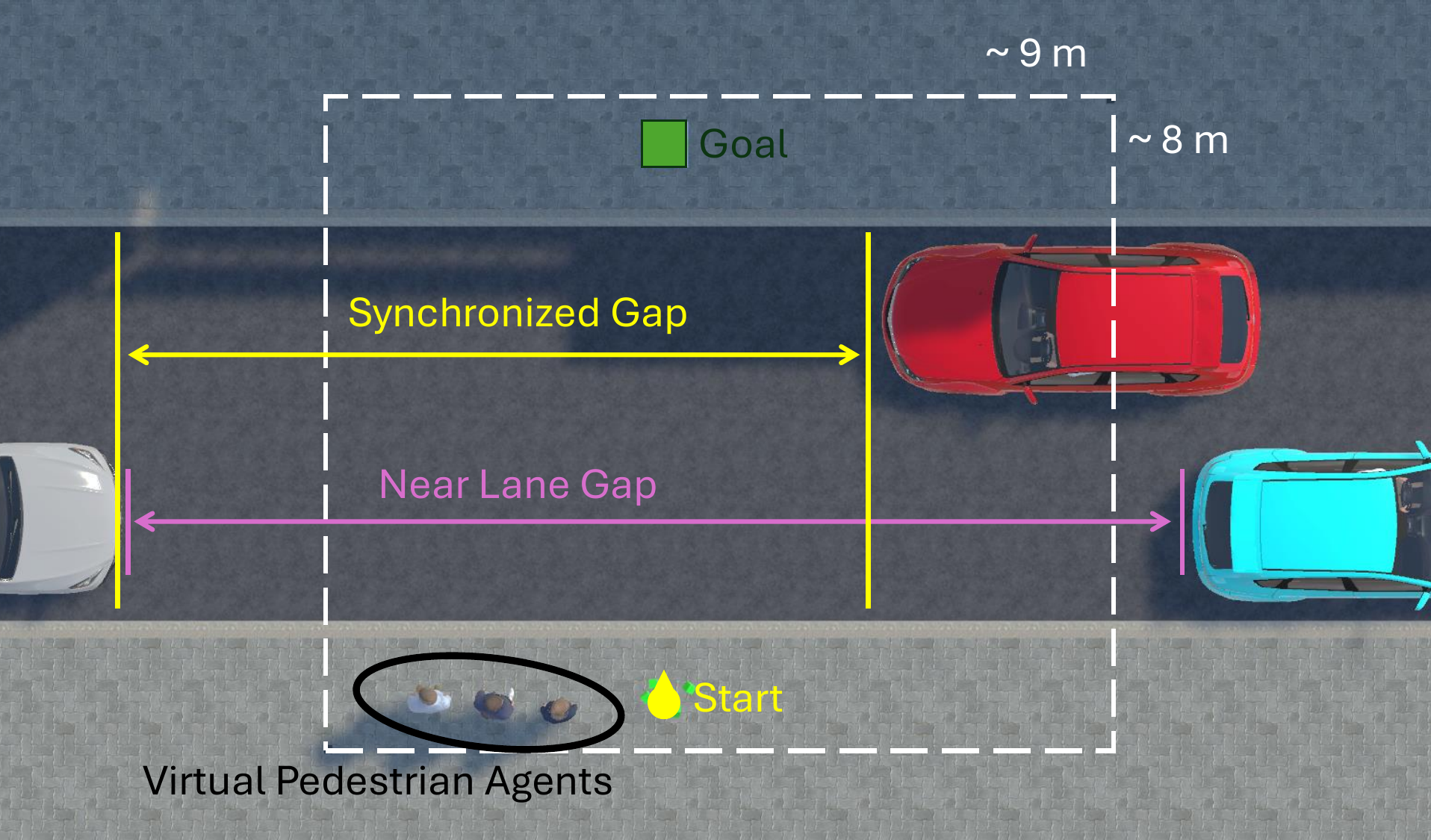}
        \label{fig:schematicPedestrians}}
    \hfill
    \subfloat[Zebra environment]{        \includegraphics[page=3,width=0.47\textwidth,trim=20 0 20 0,clip]{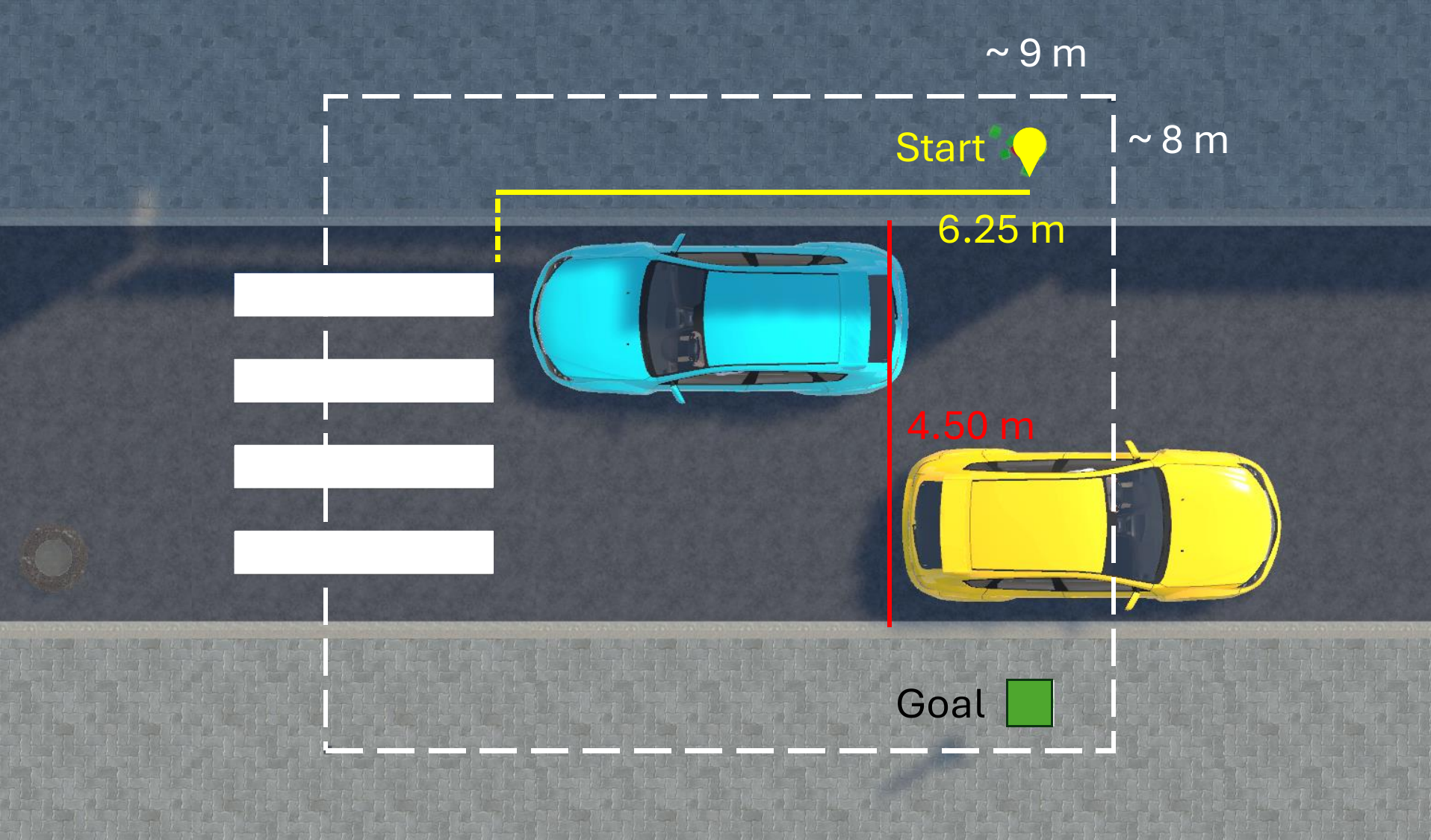}
        \label{fig:schematicZebra}}
    \caption{Schematic overview of the experimental environment. Start (yellow) and goal (green) are visualized and alternated between north and south sides of the road in every other trial. Cars were approaching from both directions with randomly selected gaps per lane. }
    \label{fig:schematicExperiment}
\end{figure*}

\subsection{Data Collection}
We use the dataset proposed by Sprenger et al.~\cite{sprenger2023cross}, focusing specifically on the German data. The data was captured in a controlled experiment in a virtual street environment with a bi-directional one-lane road. With an untethered and head-mounted virtual reality headset, participants were able to move freely in a 9~m~x~8~m space, decide their route choice in different scenarios, and physically execute the crossing by walking and running to the virtual goal on the other side of the road. A schematic overview of the experimental setup is in Fig.~\ref{fig:schematicExperiment}. 
The data was captured after the evaluation and approval by the ethical review committee.

Sixty participants, 30 of whom were females, with a mean age of 25.42 years, (standard deviation = 5.93, range = 20 - 50), were included in the study.
For each participant, 15 trials without any crossing facility, 15 trials with a zebra crossing, and 30 trials with different combinations of risky and safely crossing virtual pedestrian avatars without zebra crossings were conducted. Within the trials, vehicles were driving a constant 30 km/h and only stopped at the zebra crossing if the participant was within its vicinity. The gaps between cars were uniformly sampled between 2.5 and 8.5~seconds.
From the experiment recordings, different parameters describing pedestrian behavior can be computed, like the average crossing velocity and the selected gap. This study focuses on the gap-related measures as described in Table~\ref{tab:input features}. While~\cite{sprenger2023cross} conducted, measured, and published the anonymized data on Github, they did not further analyze the interdependencies of variables or their predictive capability. 

\subsection{Model Inputs and Outputs}
\label{sec:model inputs and outputs}
We investigate pedestrian crossing behavior at unsignalized crossings. 
For the non-zebra scenario, we predict the combined gap (in seconds) of cars from both lanes that pedestrians select and accept for crossing. This predictive analysis extends to both individual crossings and following behavior.
For the zebra crossing scenario, we aim to predict if pedestrians choose to use the zebra crossings.
The prediction is based on information available before the pedestrian initiates the crossing. The input features are described in Table~\ref{tab:input features}.


We utilize two different gap computations as possible input features, each for three different views. Gap sizes are trivial to compute for single-lane direct crossings without any lateral movement. However, for two opposing lanes and complex pedestrian movements, different measures can be utilized. First, we can either compute the gap in traffic from the perspective of the vehicles, by computing the temporal distance between two vehicles (\textit{car gap}); or, we can incorporate the ego-movement of the participant and observe the \textit{effective gaps} by utilizing an automated stopwatch calculating the time between a first car passing and a second one arriving at the respective spatio-temporal position of the participant. Second, gaps can be computed for each lane separately, or synchronized for both lanes.
For example, two cars are passing each other in front of the participant and there is a gap with a car 12~s away on the near lane and 6~s away on the far lane. In this case, the near lane gap is 12~s and the far lane and the synchronized gap are 6~s. 
A visualization of these gaps can be found in Fig. \ref{fig:schematicPedestrians}. All calculations are performed on time gaps in seconds, but due to the constant speed of the vehicles, a conversion to a distance in meters is trivial.

\begin{table}[]
    \caption{Input Variables}
    \centering
    \begin{tabular}{m{1cm}|m{6.8cm}}
    \hline
        Variable \newline (unit) & Description (Data type) \\
        \hline
        $T_w (s)$ & Pedestrian waiting time before crossing (continuous) \\
        $v_p (m/s)$ & Pedestrian average walking speed (continuous) \\ \hdashline
        $N_{en}$ & Number of unused effective gaps at near lane (discrete) \\
        $N_{ef}$ & Number of unused effective gaps at far lane (discrete) \\
        $N_{eb}$ & Number of unused effective gaps for both lanes (discrete) \\
        $M_{en} (s)$ & Largest missed effective gap at near lane (continuous) \\
        $M_{ef} (s)$ & Largest missed effective gap at far lane (continuous)\\
        $M_{eb} (s)$ & Largest missed effective gap for both lanes (continuous)\\ \hdashline
        $N_{cn}$ & Number of unused car gaps at near lane (discrete) \\
        $N_{cf}$ & Number of unused car gaps at far lane (discrete) \\
        $N_{cb}$ & Number of unused car gaps for both lanes (discrete) \\
        $M_{cn} (s)$ & Largest missed car gap at near lane (continuous) \\
        $M_{cf} (s)$ & Largest missed car gap at far lane (continuous) \\
        $M_{cb} (s)$ & Largest missed car gap for both lanes (continuous) \\   \hline 
    \end{tabular}
    \label{tab:input features}
\end{table}

\subsection{Gap Selection Prediction}

\paragraph{Predictive models}
The prediction of the accepted gap is a regression problem.
The following machine learning models are proposed and evaluated.
\begin{itemize}
    \item \textbf{Linear regression:} The model considers the linear dependencies between input and output variables.
    \item \textbf{Random Forest (RF) regression:} The model consists of a large number of regression trees and outputs the average prediction. Here we use 100 estimators with a maximum depth of five.
    \item \textbf{Neural network (NN):} We use a fully connected neural network that contains an input layer, an output layer, and two hidden layers. To avoid over-fitting, we use two and four nodes in each hidden layer, respectively.
\end{itemize}

\paragraph{Evaluation metrics}
We evaluate models with mean absolute error (MAE) and root mean squared error (RMSE) as defined in Eqs.~\ref{eq_mae} and~\ref{eq_rmse}, where $y_i$ is the ground truth for the $i^{th}$ trial, and $\hat y_i$ is the prediction, $n$ is the number of trials.

\begin{equation}
MAE =\frac{\Sigma|\hat y_i - y_i|}{n}
\label{eq_mae}
\end{equation}

\begin{equation}
RMSE =\sqrt{\frac{\Sigma(\hat y_i - y_i)^2}{n}}
\label{eq_rmse}
\end{equation}

\subsection{Zebra Crossing Usage Prediction}

\paragraph{Predictive models}
Predicting if pedestrians use the zebra crossing is a classification problem. The following machine learning models are proposed and evaluated:

\begin{itemize}
    \item \textbf{Logistic regression:} The model predicts the probability of an event by modeling the log-odds for the event as a linear combination of independent input variables.
    \item \textbf{Support Vector Machine (SVM):} The model aims to identify a hyperplane within the feature space for classification. The linear kernel is used here.
    \item \textbf{RF classification:} The model consists of decision trees and outputs the most selected label. Here we use 100 estimators with a maximum depth of five.
    \item \textbf{NN model:} We use a fully connected neural network that contains an input layer, an output layer, and two hidden layers. We use eight and four hidden nodes in each hidden layer, respectively.

\end{itemize}

\paragraph{Evaluation metrics}
We evaluate models with prediction accuracy (ACC) and F1 score as defined in Eqs.~\ref{eq_acc} and~\ref{eq_f1}, where P and N are the numbers of positives and negatives. TP, TN, FP, and FN are the numbers of true positives, true negatives, false positives, and false negatives, respectively.

\begin{equation}
ACC=\frac{TP+TN}{P+N}
\label{eq_acc}
\end{equation}

\begin{equation}
F1 =\frac{2TP}{2TP+FP+FN}
\label{eq_f1}
\end{equation}

\subsection{Implementation Details}
A total of 3585 trials were collected, comprising 894 trials at non-zebra crossings, 891 trials at zebra crossings, and 1800 trials for following behavior. We use five-fold cross-validation for evaluation. The dataset is randomly divided into five sets based on trials. Each time we use one set for test and the remaining for training, ensuring no overlap between training and test data. The reported performance metrics are averaged over five test sets.
Given the different scales and types of input variables, we normalize them into standard normal distributions for better convergence and stability.

\section{Results and Discussions}
\subsection{Gap Selection Prediction}
\label{sec: gap selection}
\begin{figure}[h]
    \centering
    \includegraphics{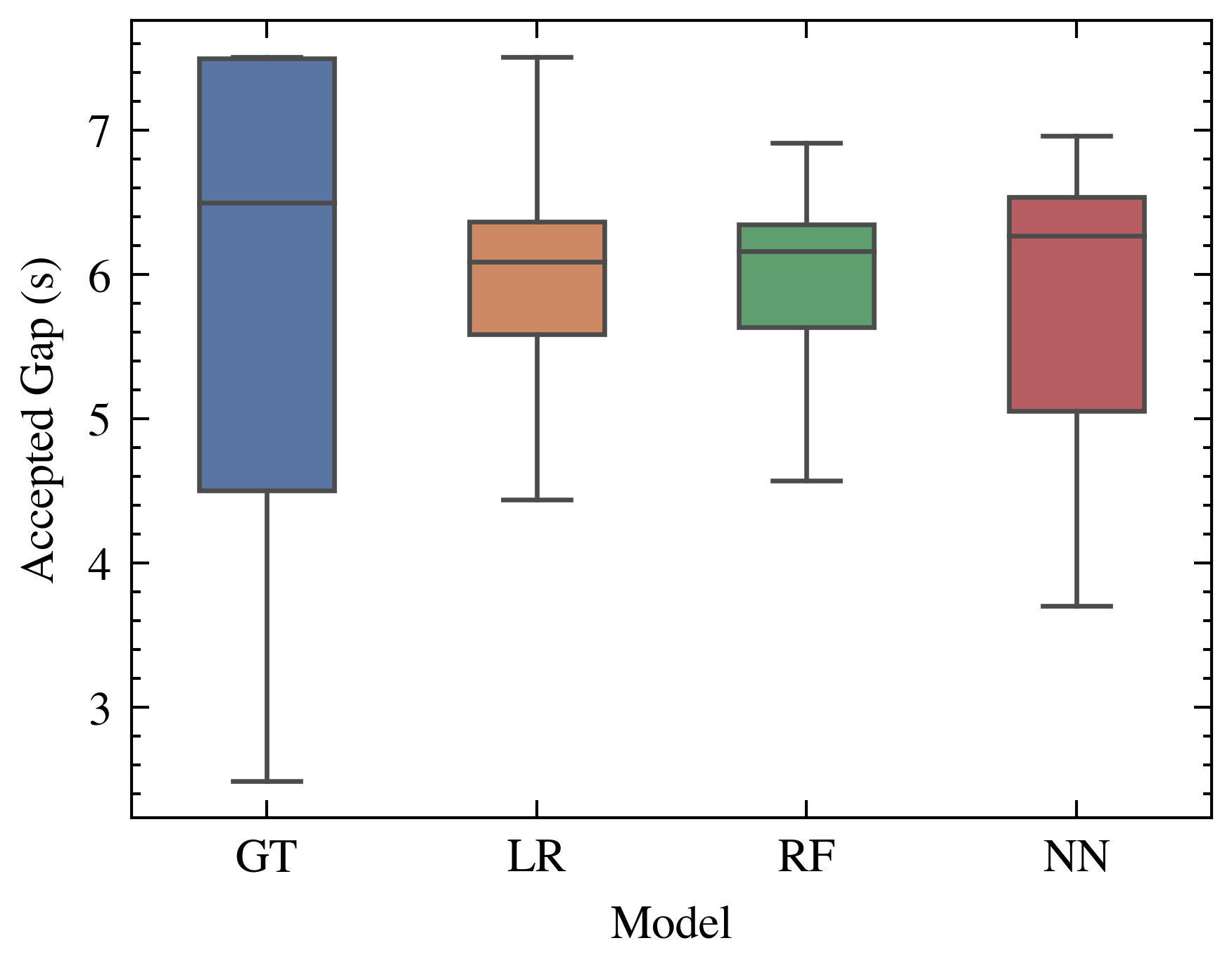}
    \caption{The boxplots of the predicted accepted gaps for different models. GT for ground truth, LR for linear regression, RF for random forest, and NN for neural network.}
    \label{fig:prediction_selected_gaps_models_boxplot}
\end{figure}

\paragraph{Quantitative results}
We predict the accepted gap for pedestrians when they cross alone at non-zebra crossings. The MAE and RMSE errors of the predicted accepted gap are detailed in Table~\ref{tab:regression_direct_alone}. The NN model outperforms other models with an MAE of 1.07 seconds. In Fig.~\ref{fig:prediction_selected_gaps_models_boxplot}, we present the boxplots illustrating the predicted results of three models.
Prediction results of linear regression and random forest models exhibit a more concentrated distribution, whereas the neural network results are more dispersed, resembling the distribution pattern of the ground truth.

\begin{table}[h]
    \centering
    \caption{The prediction error of gap selection. Unit: seconds. $T_w$ is pedestrian waiting time, $v_p$ is the pedestrian walking speed, $N_{cb}$ is the number of unused car gaps for both lanes, $N_{ef}$ is the number of unused effective gaps at far lane, $M_{cb}$ is the largest missed car gap for both lanes.}
    \label{tab:regression_direct_alone}
    \begin{tabular}{c|c|c|l}
    \hline
        Model & MAE & RMSE & Three most important features \\ \hline \hline
        Linear & 1.09 & 1.34 & $v_p, N_{cb}, N_{ef}$ \\ \hline
        RF & 1.09 & 1.35 & $M_{cb}, v_p, T_w$\\ \hline
        NN & 1.07 & 1.33 & $N_{cb}, T_w, M_{cb}$\\
        \hline
    \end{tabular}
\end{table}

The three most important features of each model are shown in Table~\ref{tab:regression_direct_alone}. For the Linear model, feature importance is determined by the coefficients in the regression function. 
The RF model computes feature importance by the mean and standard deviation of impurity decrease across all trees. For the NN model, feature importance is determined using permutation importance.

The number of unused car gaps for both lanes ($N_{cb}$), the largest missed car gap for both lanes ($M_{cb}$), the pedestrian waiting time before crossing ($T_w$), and the pedestrian average walking speed ($v_p$) are important features for more than one model. Therefore, we analyze the impact of these features on pedestrian behavior.

\begin{figure}
    \centering
    \includegraphics{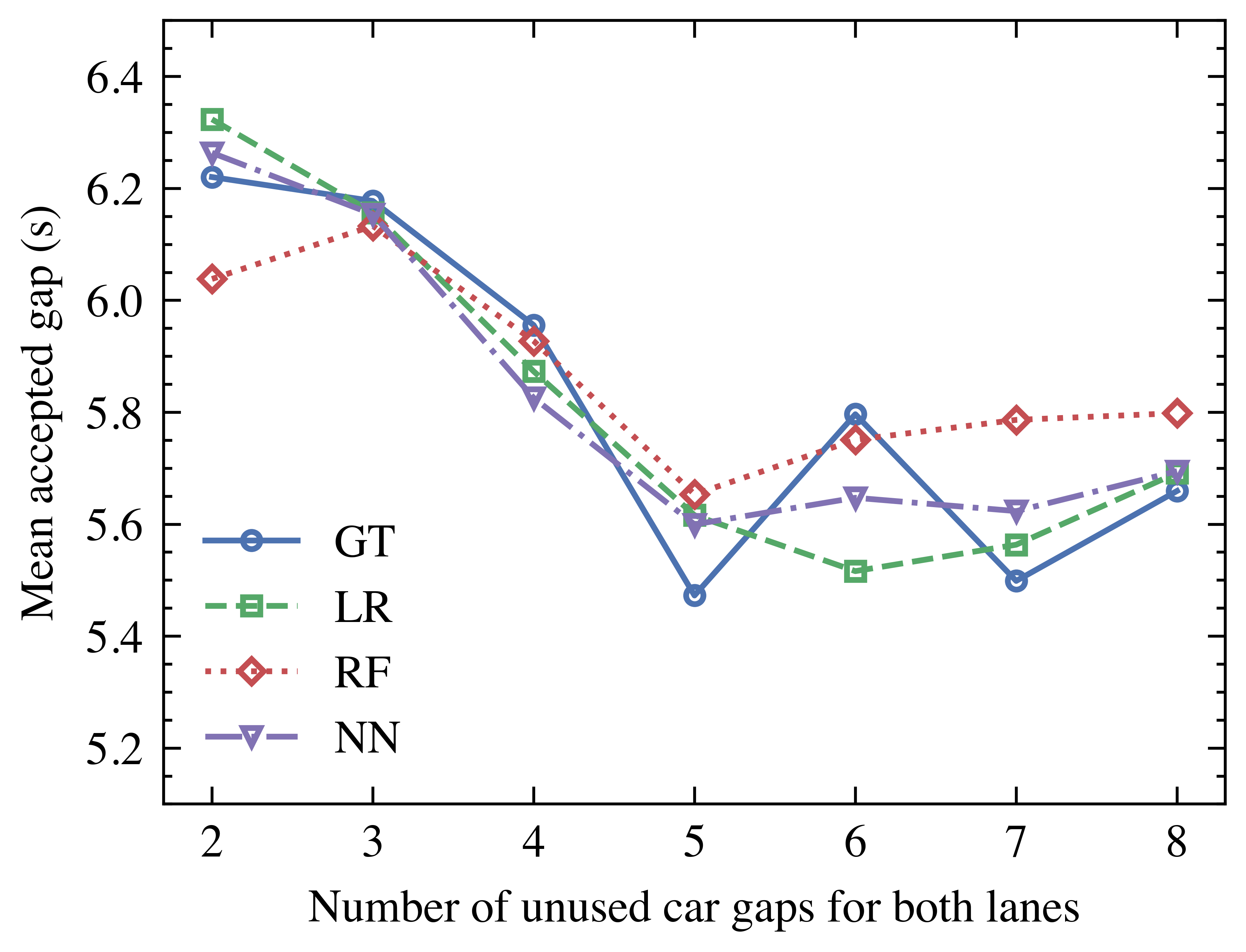}
    \caption{The mean value of the accepted gap versus the number of unused car gaps for both lanes. Pedestrians tend to accept smaller gaps for crossing as the number of unused car gaps increases.}
    \label{fig:prediction_accepted_gap_mean_vs_unused_sync_cargap_num}
\end{figure}

\paragraph{Number of unused car gaps}
Fig.~\ref{fig:prediction_accepted_gap_mean_vs_unused_sync_cargap_num} illustrates the relationship between mean accepted gaps and the number of unused car gaps for both lanes.
As the number of unused car gaps increases, the size of accepted gaps tends to decrease. This observation indicates that when pedestrians miss more gaps, there is an inclination toward making riskier choices. However, when the number of missed gaps exceeds five, the accepted gaps no longer decrease and stabilize between 5.4~s to 5.8~s. This indicates the existence of a safety threshold for accepted gaps. Pedestrians may feel safe to cross when the gap exceeds this threshold. Conversely, if a gap falls below this threshold, people may perceive it as unsafe and choose not to cross.  

\begin{figure}[tb]
    \centering
    \includegraphics{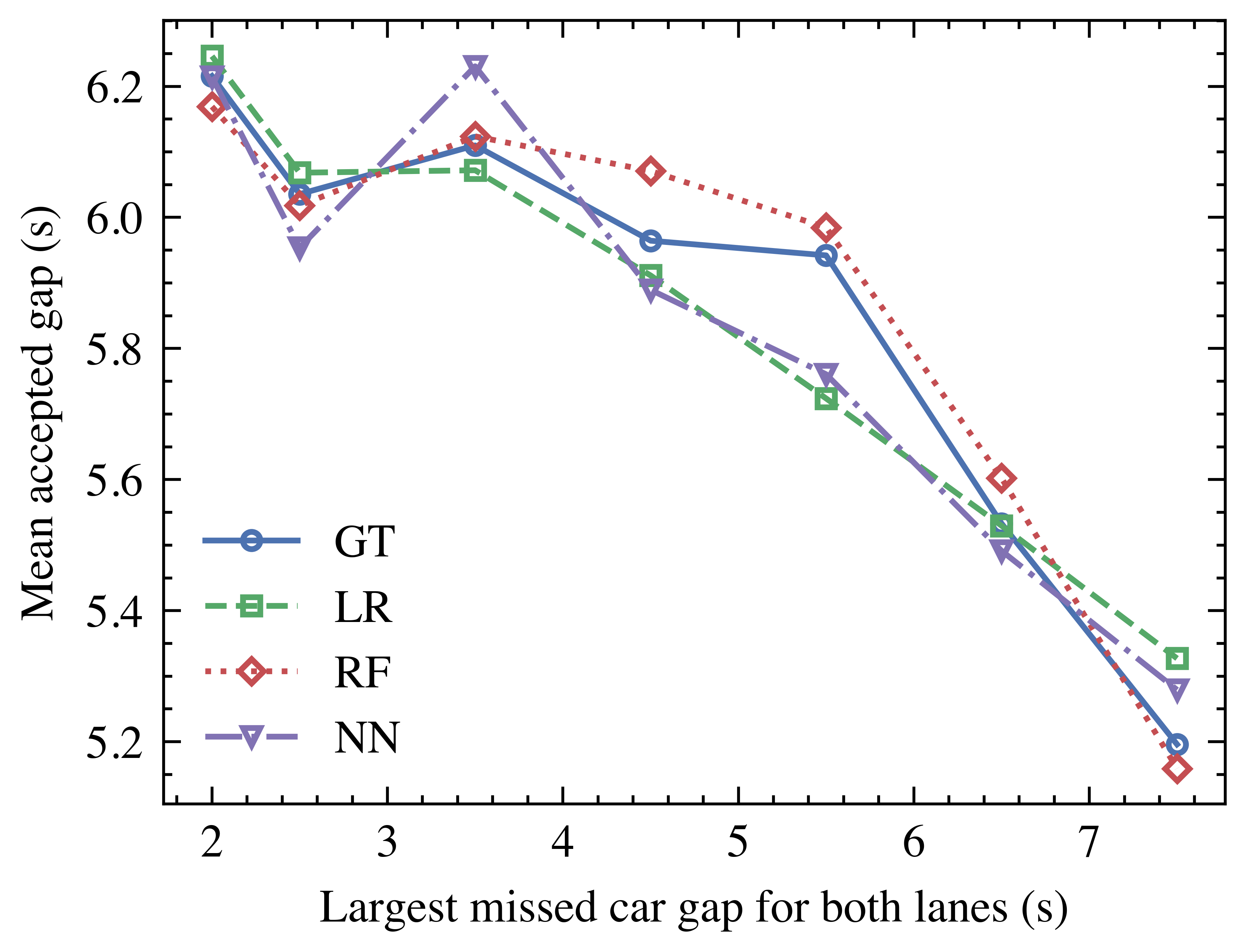}
    \caption{The mean value of the accepted gap versus the largest missed car gap for both lanes. Pedestrians tend to accept smaller gaps for crossing as the largest missed gap increases.}
    \label{fig:prediction_accepted_gap_mean_vs_largest_missed_sync_gap}
\end{figure}

\paragraph{Largest missed car gap}
The relationship between the mean accepted gap and the largest missed car gap for both lanes is shown in Fig.~\ref{fig:prediction_accepted_gap_mean_vs_largest_missed_sync_gap}. It is shown that as pedestrians miss larger gaps, they are inclined to choose smaller gaps for crossing. This indicates that pedestrians tend to make riskier choices when missing crossing chances with larger gaps.

\begin{figure}[tb]
    \centering
    \includegraphics{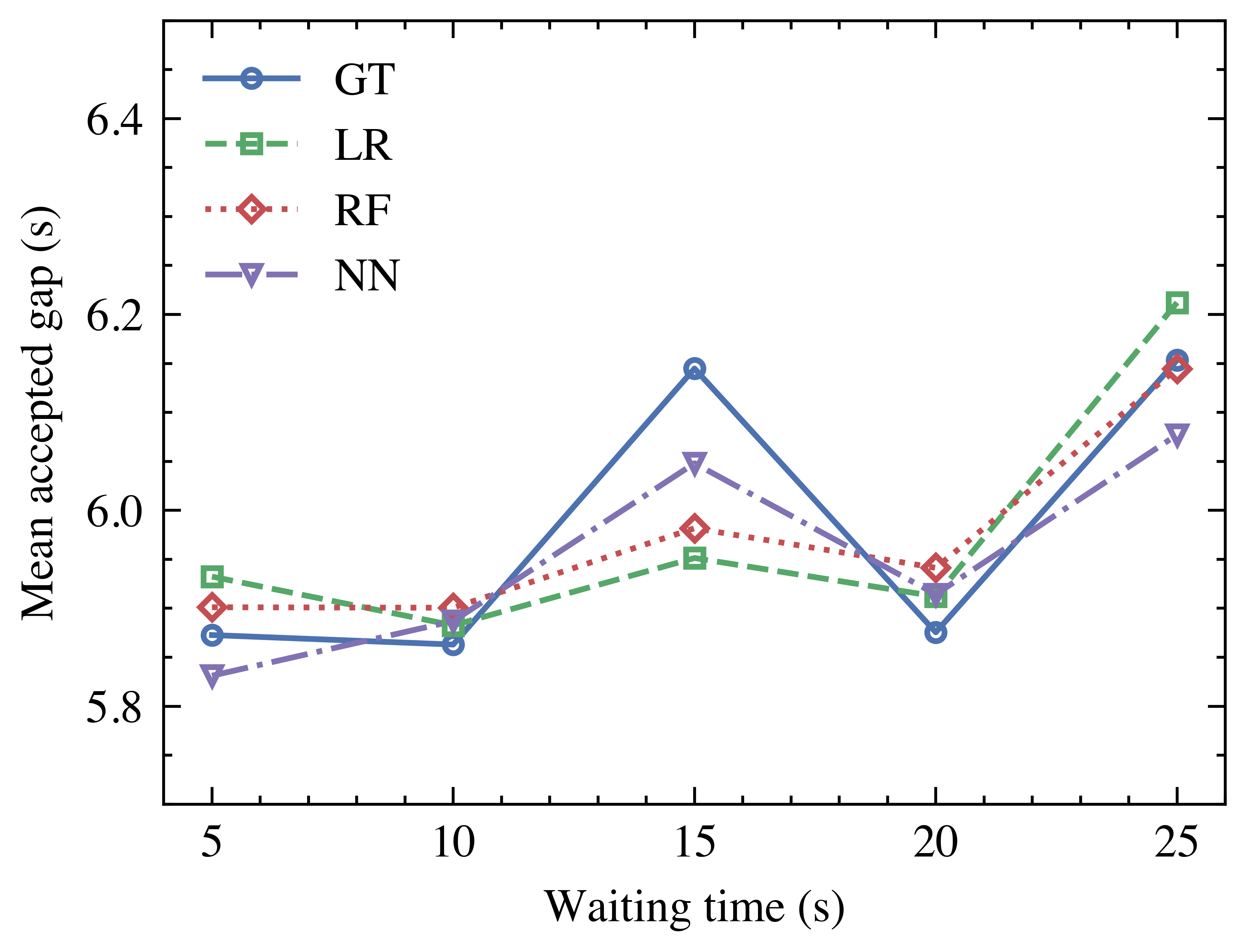}
    \caption{The mean value of the accepted gap versus pedestrian waiting time. Pedestrians tend to select larger gaps when the waiting time increases.}
    \label{fig:prediction_accepted_gap_mean_vs_waiting_time}
\end{figure}

\paragraph{Pedestrian waiting time}
The relationship between the accepted gap and waiting time is illustrated in Fig.~\ref{fig:prediction_accepted_gap_mean_vs_waiting_time}. 
While the observation indicates that pedestrians tend to make riskier choices when missing more and larger gaps, the converse is true for waiting time. Pedestrians tend to select larger gaps when they wait longer. This finding is consistent with the results obtained by Yannis et al.~\cite{yannis2013pedestrian}, suggesting that as pedestrians wait longer to cross the street, the probability of crossing decreases.
Although this might appear counter-intuitive, it can be explained by considering that pedestrians intending to wait for a longer time are generally more cautious and less likely to take risks. Therefore, those who decide to wait longer tend to choose safer gaps when they eventually decide to cross.

\begin{figure}[tb]
    \centering
    \includegraphics{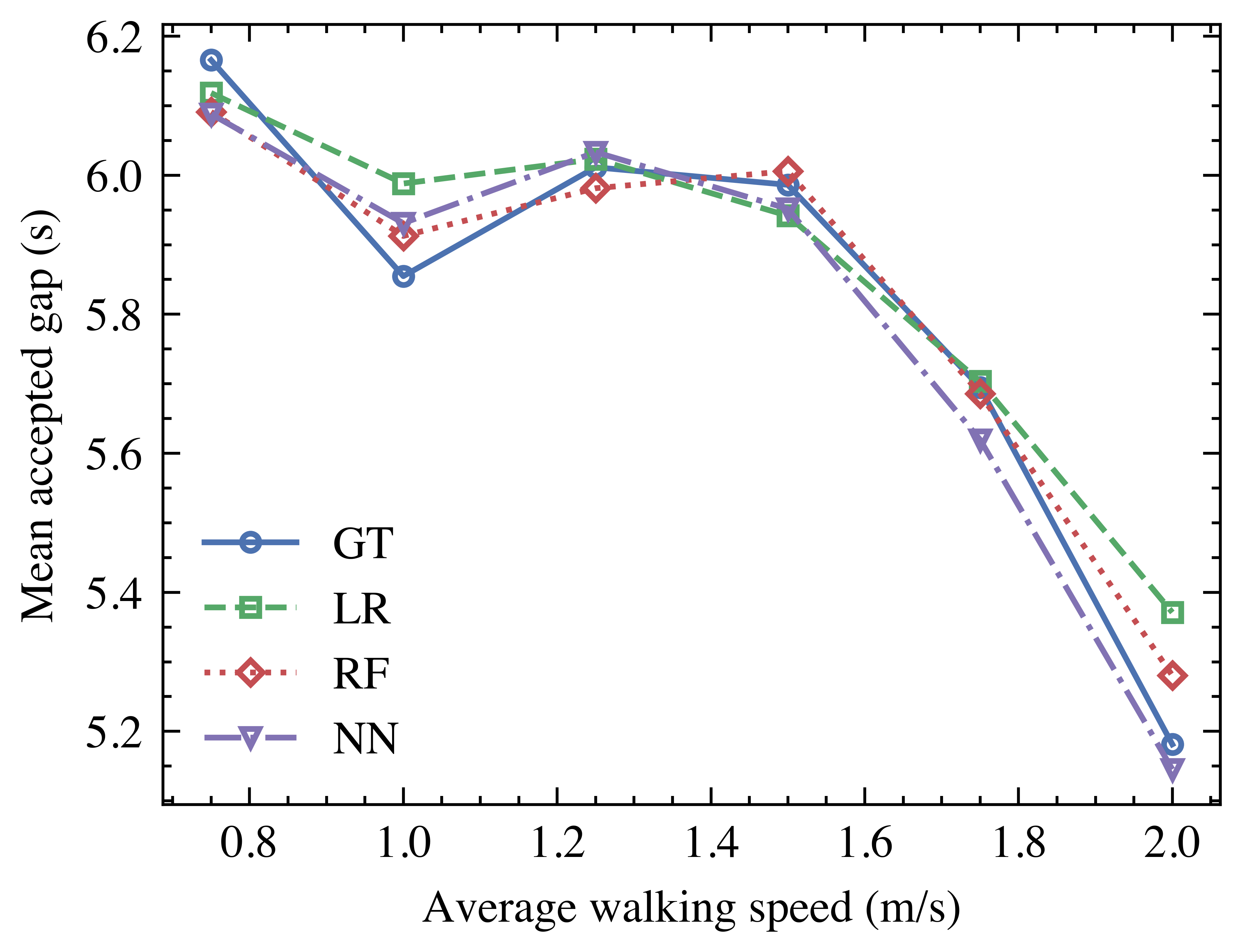}
    \caption{The mean value of the accepted gap versus pedestrian walking speed. Pedestrians tend to accept smaller gaps as the walking speed increases.}
    \label{fig:prediction_accepted_gap_mean_vs_velocity}
\end{figure}

\paragraph{Average walking speed}
The relationship between the accepted gap and pedestrian average walking speed is illustrated in Fig.~\ref{fig:prediction_accepted_gap_mean_vs_velocity}. 
The results indicate that pedestrians with faster walking speeds are inclined to choose shorter gaps for crossing. This finding is consistent with the study by Wan and Rouphail~\cite{wan2004simulation}. They proposed the critical gap formula for pedestrian crossing behavior, as defined in Eq.~\ref{eq_cg}, where L is the crosswalk length, S is the pedestrian's average walking speed, and F is a safety margin (in seconds) reflecting the pedestrian's risk acceptance. The equation shows that as walking speed increases, the critical gap decreases.

\begin{equation}
\label{eq_cg}
    Gap = L/S + F
\end{equation}

In an alternative interpretation, this relationship can also be explained by considering that individuals in a hurry often prefer shorter gaps, leading to walking at a faster speed.

\paragraph{Following Behavior}

\begin{table}[b]
\caption{The prediction error of gap selection for crossing alone, risky group, and safe group. Unit: seconds}
\label{tab:group behavior}
\begin{tabular}{c|cc|cc|cc}
\hline
 & \multicolumn{2}{c|}{Alone} & \multicolumn{2}{c|}{Risky Group} & \multicolumn{2}{c}{Safe Group} \\
 \hline
Model & MAE & RMSE & MAE & RMSE & MAE & RMSE \\ \hline
Linear & 1.09 & 1.34 & 1.13 & 1.43 & 0.32 & 0.75 \\
RF & 1.09 & 1.35 & 0.90 & 1.24 & 0.24 & 0.63 \\
NN & 1.07 & 1.33 & 0.89 & 1.25 & 0.30 & 0.78 \\
\hline
\end{tabular}
\end{table}

In addition to individual crossings, we also investigate and predict group behavior. The leading pedestrians were simulated by virtual avatars. Specifically, we predict two types of group behavior: the risky group, where the leading agent crosses at a gap of 4~s, and the safe group, where the leading agent crosses at a gap of 6.5~s. This experiment aims to investigate whether pedestrians' gap acceptance is influenced by leading agents, indicating their following behavior.
The quantitative prediction results are presented in Table~\ref{tab:group behavior}. In the risky group, only the linear model exhibits larger errors, while the two non-linear models demonstrate smaller errors. This suggests an increase in non-linearity in pedestrian behavior within the risky group. In the safe group, errors are significantly smaller, indicating that pedestrian behavior is more predictable.

\begin{figure}
    \centering
    \includegraphics{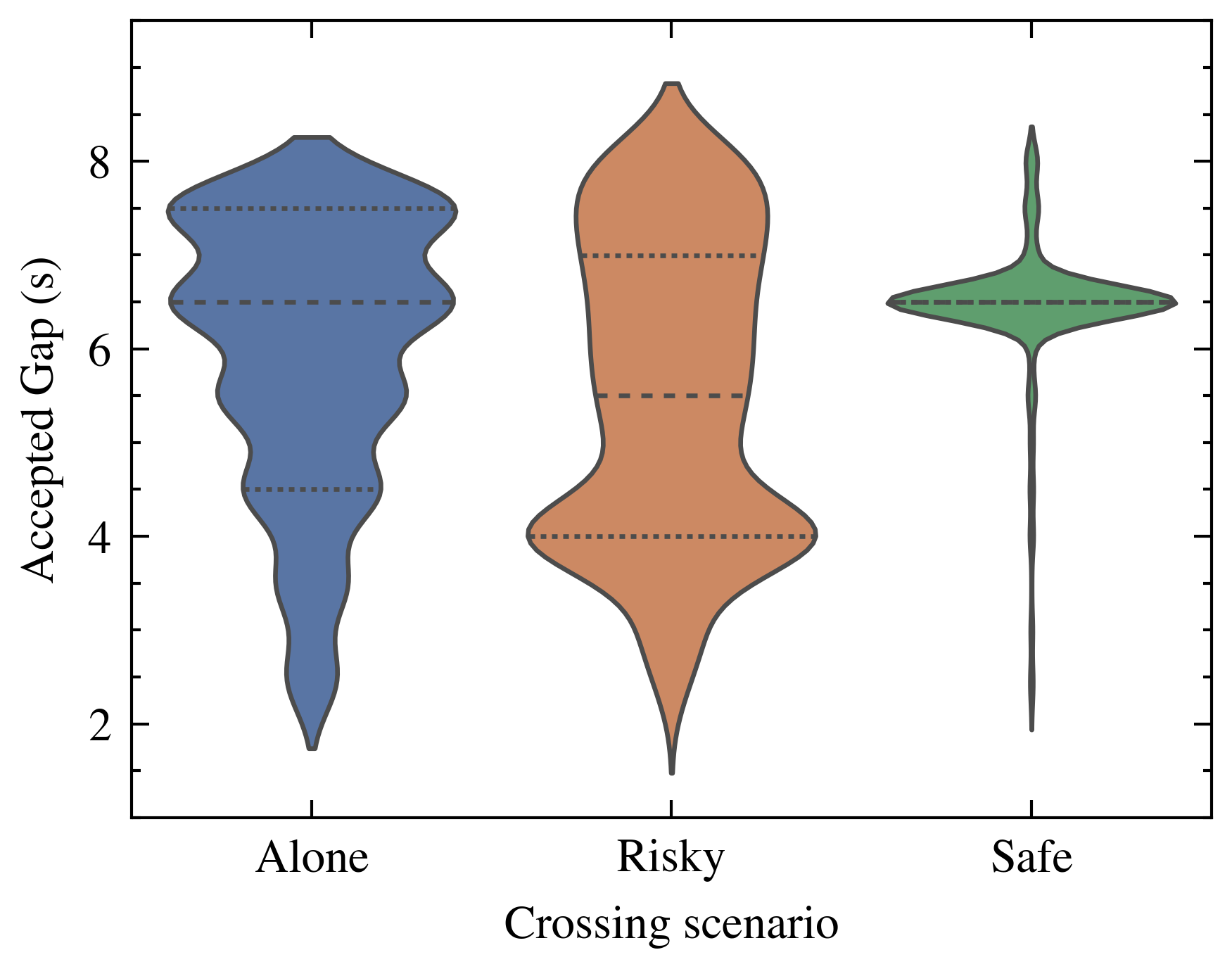}
    \caption{The distribution of accepted gaps for cross alone, the risky group, and the safe group.}
    \label{fig:compare_alone_risky_safe_group}
\end{figure}

We further look into the distributions of accepted gaps to find illustrations for the predictability of group behavior.
In consistency with Sprenger et al.'s findings~\cite{sprenger2023cross}, there are significant differences in the accepted gap among the three groups (ANOVA F(2)=107.5, p<0.001). In addition to the statistical hypothesis testing, we observe and compare the distributions of three groups. The distributions of three scenarios are shown in Fig.~\ref{fig:compare_alone_risky_safe_group}. The observation reveals that pedestrians tend to follow the behavior of leading pedestrians.
Compared with the scenario of crossing alone, the distributions of accepted gaps for both the safe and risky groups demonstrate a noticeable shift towards the selected gap of the leading pedestrian. For the risky group, while many people keep their safer choices, a considerable number of pedestrians choose to cross following the risky leading pedestrian at a 4~s gap. For the safe group, most people choose to cross following the leading pedestrian at 6.5~s.

\subsection{Zebra Crossing Usage Prediction}
\paragraph{Quantitative results}
For pedestrians near the zebra crossing, we predict whether they use zebra crossings. The accuracy and F1 score of the prediction results are presented in Table~\ref{tab:accuracy_zebra_ussage}. The NN model outperforms the other models, achieving an accuracy rate of 94.27\%. In comparison to logistic regression, a commonly employed model in analysis, the NN model exhibits a 4.02\% better accuracy and a 4.68\% better F1 score, showing the effectiveness of the proposed neural network model.

\begin{table}[h]
    \centering
    \caption{The prediction accuracy and F1 score for the use of zebra. $T_w$ is the pedestrian waiting time, $v_p$ is the pedestrian walking speed, $N_{en}$ is the number of unused effective gaps at the near lane, $N_{cn}$ is the number of unused car gaps at near lane, $M_{cn}$ is the largest missed car gap at the near lane, $M_{ef}$ is the largest missed effective gap at the far lane, $M_{eb}$ is the largest missed effective gap for both lanes, $ M_{en}$ is the largest missed effective gap at the near lane.
    }
    \label{tab:accuracy_zebra_ussage}
    \begin{tabular}{c|c|c|m{3.5cm}}
    \hline
        Model & ACC (\%) & F1 (\%) & Three most important features \\ \hline \hline
        Logistic & 90.25 & 89.22 & $N_{en}, v_p, M_{cn}$ \\ \hline
        SVM & 91.74 & 91.08 & $N_{en}, N_{cn}, M_{cn}$\\ \hline
        RF & 91.85 & 91.73 & $T_w, M_{ef}, M_{eb}$\\ \hline
        NN & 94.27 & 93.91 & $T_w, M_{en}, M_{cn}$\\
        \hline
    \end{tabular}
\end{table}
        
The three most important features in modeling are outlined in Table~\ref{tab:accuracy_zebra_ussage}. Linear-based models, including Logistic Regression and SVM, rely more on features related to the near lane. Both models identify the number of unused effective gaps at near lane ($N_{en}$) as the most important feature. In contrast, non-linear models, including RF and NN, prioritize pedestrian waiting time ($T_w$) as the most important feature for modeling. Therefore, we look into the impact of these two factors on prediction accuracy.

\begin{figure}
    \centering
    \includegraphics{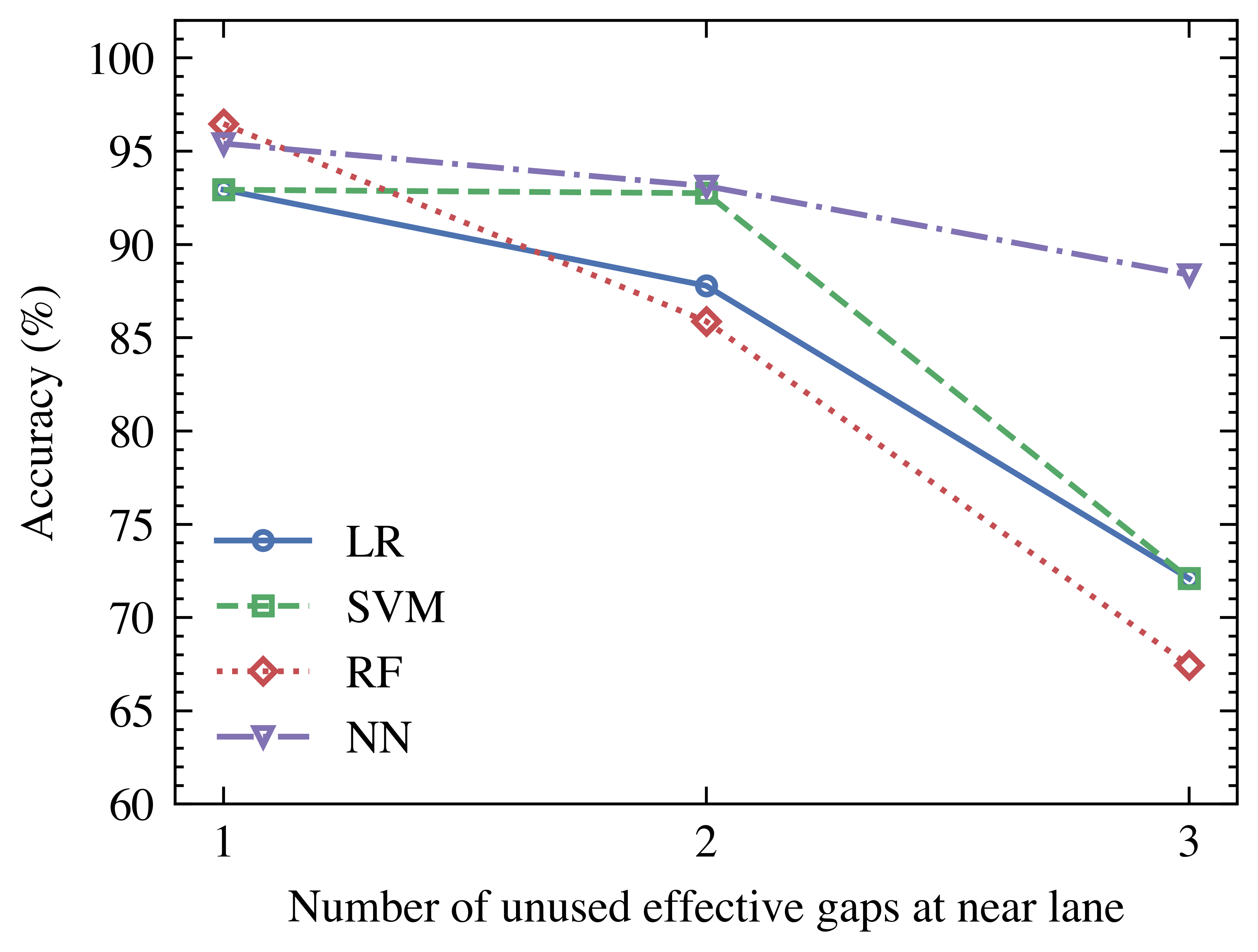}
    \caption{The prediction accuracy versus the number of unused effective gaps at near lane.}
    \label{fig:prediction_accuracy_vs_eff_near_gap_num}
\end{figure}

\paragraph{Number of unused effective gaps}
The relationship between prediction accuracy and the number of unused effective gaps at the near lane is presented in Fig.~\ref{fig:prediction_accuracy_vs_eff_near_gap_num}. It is shown that as the number of unused effective gaps at the near lane increases, prediction accuracy decreases, indicating the increased difficulty in predicting zebra usage. It is noteworthy that neural networks consistently exhibit high accuracy across all gap numbers compared to other models, demonstrating its capability of prediction.

\begin{figure}
    \centering
    \includegraphics{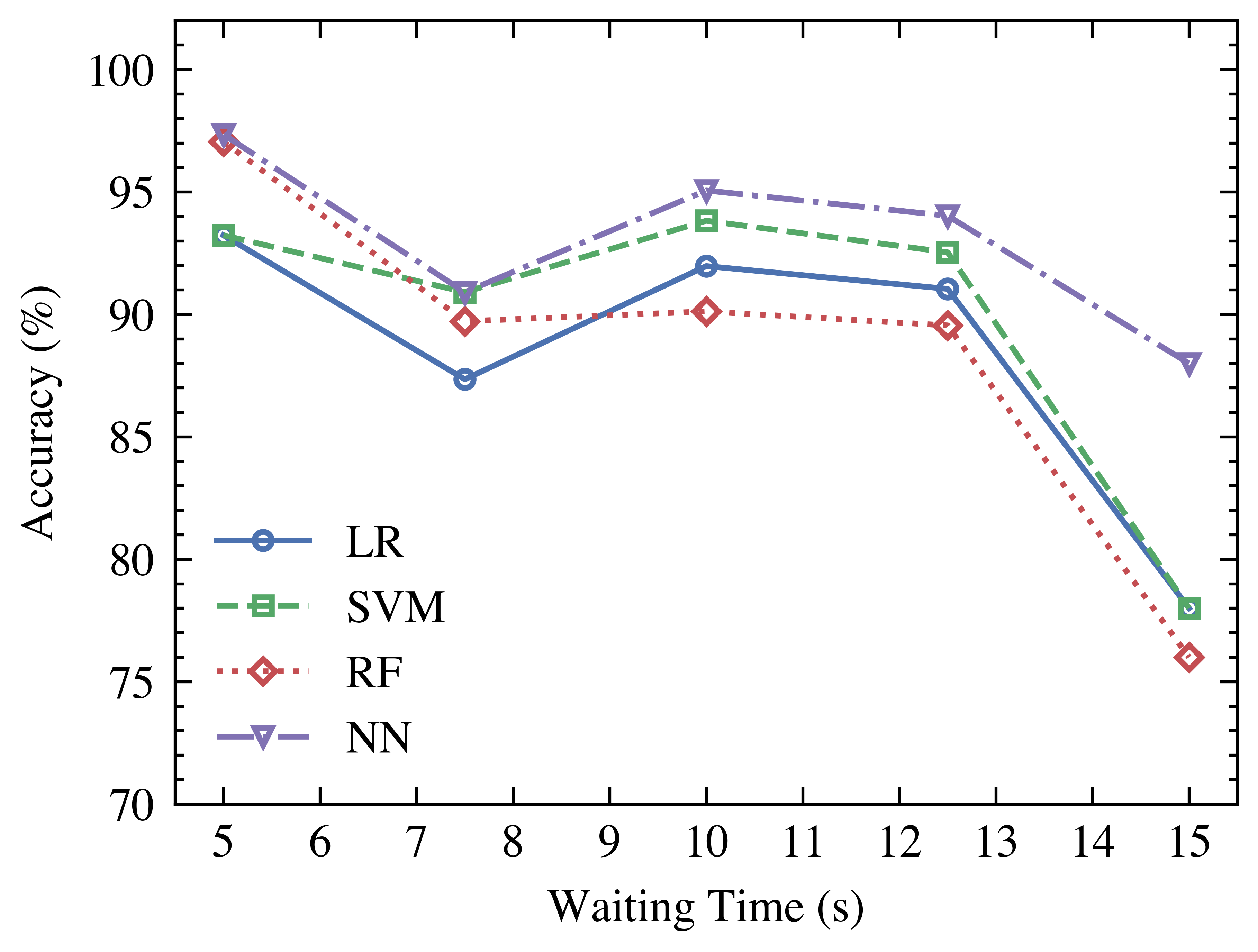}
    \caption{The prediction accuracy versus pedestrian waiting time.}
    \label{fig:prediction_accuracy_vs_waiting_time}
\end{figure}

\paragraph{Pedestrian waiting time}
The relationship between prediction accuracy and pedestrian waiting time is presented in Fig.~\ref{fig:prediction_accuracy_vs_waiting_time}. It is shown that, with increasing waiting time, the predictability of zebra crossing usage decreases. This suggests that as pedestrians wait longer, their decisions become more influenced by implicit factors, making them harder to predict. Meanwhile, the neural network consistently exhibits better results compared to other models.

\begin{figure}
    \centering
    \includegraphics{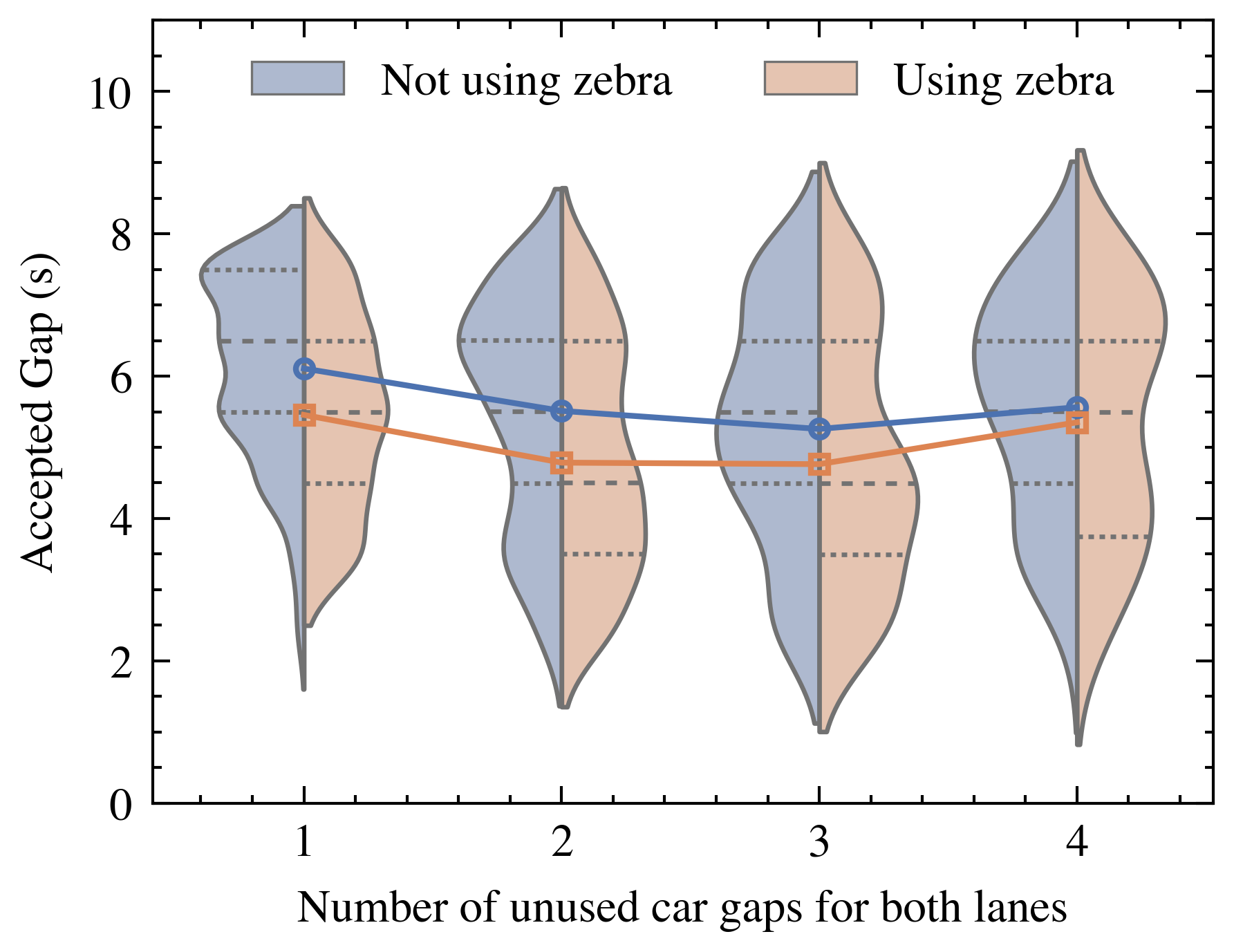}
    \caption{The distributions of the accepted gaps for pedestrians who used and did not use zebra crossings in relation to the number of unused car gaps for both lanes. Violin plots show quartiles representing the 25th (top line), 50th (middle line), and 75th (bottom line) percentiles of the distribution. Solid lines indicate the mean value of the distribution.}
    \label{fig:accepted_gap_vs_zebra_using_vs_sync_unused_cargap}
\end{figure}

\paragraph{Accepted gaps}
We compare the distribution of accepted gaps between pedestrians using and not using the zebra crossing, along with the relationship with the number of unused car gaps for both lanes, as shown in Fig.~\ref{fig:accepted_gap_vs_zebra_using_vs_sync_unused_cargap}. The distribution of the accepted gap differs for pedestrians who used the zebra crossing and those who did not. For pedestrians who used the zebra crossing, the mean accepted gap is smaller than for pedestrians who did not use the zebra crossing. This indicates two possible reasons. First, pedestrians are more inclined not to use the zebra crossing when larger gaps are available. Second, pedestrians' risk aversion may decrease in the presence of the zebra crossing, as they expect the vehicles to stop or yield. 
Furthermore, among pedestrians choosing not to use the zebra crossing, a higher number of unused car gaps for both lanes corresponds to a smaller gap selected by pedestrians. This finding aligns with the results for the non-zebra crossing cases in Sec.~\ref{sec: gap selection}. However, for pedestrians choosing to use the zebra crossing, this pattern is less obvious. This might be attributed to the fact that cars will stop at the zebra crossing if there is a pedestrian crossing the street, diminishing the significance of the gap selection in the crossing behavior.

\subsection{Implications for Intelligent Vehicles}

We focus on predicting and analyzing pedestrian gap selection and zebra crosswalk usage behavior at unsignalized crossings. The proposed models have the potential to enable intelligent vehicles to predict pedestrian behavior in advance.
By providing more action time, our study contributes to a smoother and safer interaction between pedestrians and vehicles. Vehicles can adjust their speed and trajectories based on predictive information. This leads to more proactive and precautionary measures, thereby avoiding potential conflicts and collisions.

\subsection{Limitations and Future Work}
This research is based on data collected from participants in Germany, and the observed behavior may be influenced by cultural effects. Results in different cultures may vary, suggesting the need for further studies to investigate cultural differences between countries.
Besides, the predictive model in this paper treats gap duration as a continuous value, overlooking the inherent uncertainty in pedestrian gap selection. Future research can address this limitation by exploring stochastic models that take into account the uncertainty and randomness associated with pedestrian crossing behavior.
Additionally, in this study, we used the simulator study to prevent real-world near-crash scenarios and ensure pedestrian safety. Although participants received training sessions to emulate real-life behavior, the absence of real risk in the simulation may lead to riskier crossing behaviors compared to real-life scenarios.

\section{Conclusions}
In conclusion, this study contributes to the understanding and prediction of pedestrian behavior at unsignalized crossings. The proposed machine learning models demonstrate effectiveness in predicting gap selection and the use of zebra crossings. Neural networks performed the best compared to other methods we employed and evaluated. Factors influencing pedestrian gap selection, including the number of unused gaps, the largest missed gap, walking speed, waiting time, and group behavior, are comprehensively investigated. We also analyzed the impact of waiting time and the number of unused gaps on the decision of zebra usage. This research proposes and evaluates predictive models for pedestrian crossing behavior, providing valuable insights into pedestrian-vehicle interactions. The findings offer practical implications for enhancing the safety and performance of automated driving systems.

\addtolength{\textheight}{-4.5cm}   









\bibliographystyle{IEEEtran}
\bibliography{IEEEexample}

\end{document}